\documentclass[acmsmall,nonacm]{acmart}

\usepackage{amsmath,amsfonts}
\usepackage{algorithm}
\usepackage{algorithmic}
\usepackage{graphicx}
\usepackage{multirow}
\usepackage{booktabs}
\usepackage{xspace}
\usepackage{enumitem}
\usepackage{pifont}
\usepackage{subcaption}

\usepackage{xcolor}

\newcommand{\bh}[1]{\vspace{3pt}\noindent \textbf{#1} }

\newcommand{\emam}{eMamba\xspace}

\usepackage{multirow}

\setcopyright{rightsretained}
\acmJournal{TECS}
\acmYear{2025}
\acmVolume{1}
\acmNumber{1}
\acmArticle{1}
\acmMonth{9}
\acmDOI{}

\begin{document}

\title{\emam: Efficient Acceleration Framework for Mamba Models in Edge Computing}

\author{Jiyong Kim}
\email{fatica@ulsan.ac.kr}
\affiliation{
    \institution{University of Ulsan}
    \department{Department of Electrical, Electronic and Computer Engineering}
    \city{Ulsan}
    \postcode{44610}
    \country{Republic of Korea}
}

\author{Jaeho Lee}
\email{dlwogh909@ulsan.ac.kr}
\affiliation{
    \institution{University of Ulsan}
    \department{Department of Electrical, Electronic and Computer Engineering}
    \city{Ulsan}
    \postcode{44610}
    \country{Republic of Korea}
}

\author{Jiahao Lin}
\email{jlin445@wisc.edu}
\affiliation{
    \institution{University of Wisconsin-Madison}
    \department{Department of Electrical and Computer Engineering}
    \city{Madison}
    \state{WI}
    \postcode{53706}
    \country{USA}
}

\author{Alish Kanani}
\email{ahkanani@wisc.edu}
\affiliation{
    \institution{University of Wisconsin-Madison}
    \department{Department of Electrical and Computer Engineering}
    \city{Madison}
    \state{WI}
    \postcode{53706}
    \country{USA}
}

\author{Miao Sun}
\email{smiao23@wisc.edu}
\affiliation{
    \institution{University of Wisconsin-Madison}
    \department{Department of Electrical and Computer Engineering}
    \city{Madison}
    \state{WI}
    \postcode{53706}
    \country{USA}
}

\author{Umit Y. Ogras$^*$}
\email{uogras@wisc.edu}
\affiliation{
    \institution{University of Wisconsin-Madison}
    \department{Department of Electrical and Computer Engineering}
    \city{Madison}
    \state{WI}
    \postcode{53706}
    \country{USA}
}

\author{Jaehyun Park$^*$}
\email{jaehyun@ulsan.ac.kr}
\affiliation{
    \institution{University of Ulsan}
    \department{Department of Electrical, Electronic and Computer Engineering}
    \city{Ulsan}
    \postcode{44610}
    \country{Republic of Korea}
}

\thanks{$^*$Corresponding authors}

\begin{abstract}
State Space Model (SSM)-based machine learning architectures have recently gained significant attention for processing sequential data. Mamba, a recent sequence-to-sequence SSM, offers competitive accuracy with superior computational efficiency compared to state-of-the-art transformer models. While this advantage makes Mamba particularly promising for resource-constrained edge devices, no hardware acceleration frameworks are currently optimized for deploying it in such environments.
This paper presents \emam, a comprehensive end-to-end hardware acceleration framework explicitly designed for deploying Mamba models on edge platforms. 
\emam maximizes computational efficiency by replacing complex normalization layers with lightweight hardware-aware alternatives and approximating expensive operations, such as SiLU activation and exponentiation, considering the target applications. 
Then, it performs an approximation-aware neural architecture search (NAS) to tune the learnable parameters used during approximation. 
Evaluations with Fashion-MNIST, CIFAR-10, and MARS, an open-source human pose estimation dataset, show \emam achieves comparable accuracy to state-of-the-art techniques using 1.63--19.9$\times$ fewer parameters.
In addition, it generalizes well to large-scale natural language tasks, demonstrating stable perplexity across varying sequence lengths on the WikiText2 dataset.
We also quantize and implement the entire \emam pipeline on an AMD ZCU102 FPGA and ASIC using GlobalFoundries (GF) 22~nm technology.
Experimental results show 4.95--5.62$\times$ lower latency and 2.22--9.95$\times$ higher throughput, with 4.77$\times$ smaller area, 9.84$\times$ lower power, and 48.6$\times$ lower energy consumption than baseline solutions while maintaining competitive accuracy.
\end{abstract}




\maketitle

\vspace{2mm}
\section{Introduction} \label{sec:introduction}
Deep learning (DL) models are progressing at an unprecedented pace, driving breakthroughs across a wide range of domains. However, the growing complexity of recent models, particularly transformer-based architectures such as Vision Transformers (ViTs), comes at a steep cost~\cite{xia2024understanding,shekhar2024towards}. They demand substantial computational and storage resources, which require powerful GPUs that consume hundreds of watts. This results in massive energy consumption and a carbon footprint that has started impacting the power grid~\cite{de2023growing,van2021sustainable}. 
Moreover, these requirements render complex DL models impractical for edge processing, where inference can be performed with orders of magnitude lower energy footprint while preserving privacy. Hence, \textit{there is a strong need for lightweight, energy-efficient DL models for resource-constrained edge processing}.

One of the key limitations of transformers is their quadratic complexity during inference, driven by costly key-value calculations. 
State-space models (SSMs) have recently been proposed to address this problem and enable more efficient DL. SSMs are a classical mathematical framework for describing dynamic systems using first-order differential equations, with applications in control systems, signal processing, and electrical circuit design~\cite{bishop2011modern,alexander2007fundamentals}. 
They have recently gained attention in machine learning (ML) for their remarkable effectiveness in modeling sequential data~\cite{gu_Efficiently_2022, gu_How_2022}. Modern SSM-based ML architectures have demonstrated promising results, offering faster training and efficient inference, especially for long sequences~\cite{gu_Mamba_2023, wang2024graph}. Mamba is a recent SSM-based architecture that achieves competitive performance with transformers \textit{with linear time complexity}, a substantial improvement over the quadratic complexity of attention mechanisms in transformers~\cite{gu_Mamba_2023}. Moreover, prior work established its equivalence to Convolutional Neural Networks (CNNs) and Recurrent Neural Networks (RNNs), proving its versatility and applicability across various ML tasks~\cite{gu_Combining_2021}.

\begin{figure*}[b!] 
\centering
    \centering
    \vspace{-3mm}
    \includegraphics[width=0.99\linewidth]{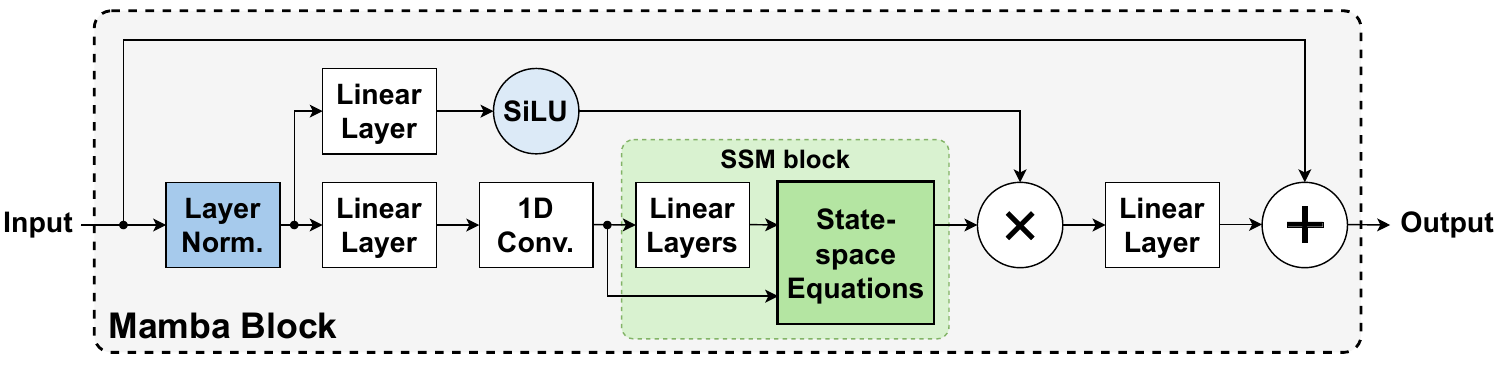}
    \vspace{-5pt}
    \caption{Overview of Mamba architecture. The highlighted blocks are described in detail.}
    \label{fig:mamba_arch}
\end{figure*}

The Mamba begins with layer normalization to ensure consistent feature scaling and stabilize training, as illustrated in Figure~\ref{fig:mamba_arch}. Then, the data is processed through two parallel linear layers that expand the feature dimension. The upper path with a SiLU activation serves as a gating mechanism, controlling the output of SSM via element-wise multiplications. A 1D convolution is applied in the lower path to mix input features, preparing them for temporal modeling. 
Then,  the SSM layer, Mamba's key feature, models the system dynamics and sequence dependencies using a discretized SSM, offering an efficient alternative to attention mechanisms. Finally, a linear projection reverts the expanded features back to the original dimension, and a residual connection is added to preserve input information and further stabilize training.

Since the key advantage of SSM and Mamba models is providing competitive performance with lower complexity, \textit{this paper focuses on their resource-constrained implementation for edge AI applications}.  
We present a novel end-to-end framework to design efficient Mamba (\emam) hardware accelerators. 
While substantial research has targeted accelerating mainstream ML models like deep neural networks and transformers~\cite{chen_Eyeriss_2016, samajdar_SCALESim_2019, you_ViTCoD_2022}, Mamba is still relatively underexplored, despite its promising potential. 
Only two recent studies present reconfigurable accelerator architectures for Mamba~\cite{li2024marca,ligthmamba_2025}. 
\textit{However, both target large language models (LLM), as discussed in the related work. 
No comprehensive frameworks exist for accelerating Mamba models, specifically in edge scenarios, where their linear time complexity could be fully leveraged.}
To address this gap, we present \emam, an end-to-end optimization framework designed to accelerate Mamba models for edge applications. We demonstrate its effectiveness on both an FPGA platform and using GlobalFoundries (GF) 22nm technology, highlighting its suitability for real-world, energy-efficient deployment.

\emam introduces a suite of hardware-friendly optimizations to tackle the unique challenges of accelerating Mamba models, including managing recurrence in SSMs, handling non-linear operations, and minimizing latency through computation unrolling. It replaces computationally expensive layer normalization and SiLU with efficient alternatives like range normalization and piecewise linear approximations, achieving substantial gains in efficiency with minimal impact on accuracy. 
All weights and activations are quantized to minimize computation and memory footprint further.
In addition, \emam uses approximation-aware training and architecture search to find cost-effective model configurations that balance accuracy and resource efficiency. It also applies custom unrolling within the pipeline to optimize the latency–resource usage trade-off.

We evaluate \emam on three representative vision datasets, Fashion-MNIST~\cite{fashionMNIST_2017}, CIFAR-10~\cite{cifar10_2009}, and MARS~\cite{an_MARS_2021}, and  compare it against recent ViT and CNN models.
Across these tasks, we achieve comparable accuracy to the baseline ViT models~\cite{learnable_vit_2023, understanding_vit_2023, ViT_iclr_2021} while reducing the model size by 19.9$\times$, 2.02$\times$, and 1.63$\times$ for Fashion-MNIST, CIFAR-10, and MARS, respectively.
In addition to small-scale vision tasks, \emam demonstrates strong generalization to complex language tasks, as reflected by its stable performance across varying sequence lengths on the WikiText2~\cite{wikitext2_2016} dataset, outperforming RNN~\cite{rnn_2010}, LSTM~\cite{lstm_2014}, and ViT models of comparable size.
We also evaluated the performance and area using AMD Vivado on the ZCU102 FPGA board and Synopsys Design Compiler \& IC Compiler II with GF 22nm technology.
Hardware measurements for the MARS datasets show that \emam achieves 5.62$\times$ lower latency and 9.95 $\times$ higher throughput compared to a state-of-the-art CNN accelerator.
Compared to a ViT accelerator, it achieves 4.95$\times$ lower latency and 2.22$\times$ higher throughput using 4.77$\times$ smaller area and 7.02M fewer gates based on GF 22~nm technology, along with 9.84$\times$ lower power and 48.6$\times$ lower energy consumption.

\vspace{1mm}
The key contributions in this work are as follows:
\begin {itemize}[leftmargin=*]
\item \textbf{An end-to-end Mamba design framework for edge AI}, featuring automated architecture exploration, hardware-algorithm co-optimization, and silicon-proven validation across FPGA and GF 22nm technology.
\item \textbf{High efficiency via application-aware and hardware-aware approximations} for exponential operations, a hybrid precision quantization strategy, and approximation-aware neural architecture search
(NAS) are employed, achieving comparable accuracy to CNN and ViT with \textbf{63$\times$} and \textbf{1.63$\times$} parameter reduction, respectively.
\item \textbf{Extensive performance validation on MARS dataset,} 
which demonstrates \textbf{5.62$\times$} lower latency and \textbf{9.95$\times$} higher throughput compared to iso-accuracy CNNs on the FPGA platform. Furthermore, \emam achieves \textbf{2.22$\times$} lower latency and \textbf{4.95$\times$} higher throughput compared to iso-accuracy ViTs, with \textbf{4.77$\times$} smaller area in our GF 22~nm technology implementation, along with \textbf{9.84$\times$} lower power and \textbf{48.6$\times$} lower energy consumption.
\end {itemize}

In the rest of this paper, Section~\ref{sec:related_work} reviews the related work. Section~\ref{sec:overview} overviews the proposed design framework, while Section~\ref{sec:design} details the complete end-to-end acceleration methodology. 
Section~\ref{sec:results} presents comprehensive experimental results, including the FPGA implementation and area comparison on GF~22nm technology.
Finally, Section~\ref{sec:conclusion} concludes the paper.
\section{related work}\label{sec:related_work}
Hardware acceleration of AI/ML models has become prevalent, driven by the limitations of conventional computing systems in handling the speed and scalability demands of modern data processing~\cite{chen_survey_2020}. 
A critical factor for efficient edge acceleration is leveraging data locality to minimize external data bandwidth.
To this end, Escher et al. implement flexible buffering in CNN accelerators to balance the bandwidth demands for weights and input data across different layers~\cite{shen_escher_2017}. 
Similarly, Choi et al. propose a scoreboard mechanism to dynamically allocate DRAM data to PEs to ensure high bandwidth utilization~\cite{choi_efficient_2022}.
FINN proposes an end-to-end solution to implement fast and scalable inference engines for quantized neural networks on FPGA~\cite{finn}.
For transformer models, OPTIMUS offers a flexible hardware architecture tailored to diverse matrix multiplication requirements, effectively reducing DRAM accesses~\cite{park_optimus_2020}. 
Similarly, ViTA targets Vision Transformer deployment in resource-constrained environments by employing a head-level coarse-grained pipeline~\cite{nag_vita_2023}.
It effectively reduces repeated off-chip accesses by segmenting the multi-head attention module into two dedicated processing units, each specialized for different computations but reusing the head data. 

Non-linear computations in DNNs, such as sigmoid and softplus, can disrupt the streamlined processing of accelerators in linear computation~\cite{geng2019hardware}.
To address this problem, Kosuge et al. propose an A-LUT approach that approximates non-linear outputs using lookup tables~\cite{kosuge_16_2021}. 
This method reduces the dependence on data exchange with host processors. 
Similarly, ReAFM proposes a reconfigurable non-linear activation function module that leverages the correlations among different non-linear activation functions to minimize hardware consumption for approximation~\cite{wu2023reafm}. 

Modern ML accelerators and processors widely adopt quantization to reduce computational cost and memory requirements~\cite{chen_survey_2020}.
Most commercial accelerators use INT8 as the default numerical precision for inference applications~\cite{reuther_ai_2022}. 
A key goal in quantization research is to minimize the accuracy loss introduced by reduced precision while still achieving a high compression rate.
To this end, HAQ~\cite{wang_haq_2019} implements flexible bit-width quantization, tailoring the bit-width for each neural network layer according to its computational intensity and redundancy.
It uses a reinforcement learning (RL) agent to optimize energy, latency, and accuracy.
Quantizing Mamba models presents unique challenges due to the presence of outliers, which are amplified by parallel scan operations and matrix multiplications. 
Two post-training quantization (PTQ) frameworks, MambaQuant~\cite{xu2025mambaquant} and Quamba~\cite{chiang2024quamba}, have been proposed to tackle this challenge.
MambaQuant employs a data-dependent Karhunen-Loève Transform (KLT) combined with Hadamard-based methods to align data distributions~\cite{chee2023quip}.
It operates in two modes: an offline mode that uses KLT-enhanced rotation and an online mode that fuses direct smoothing into the weights.
In contrast, Quamba targets the sensitivity of input activations to quantization noise and applies a percentile clipping strategy to eliminate the relatively small number of extreme outliers.


The interplay between data modality, dataset adaptation, and architectural optimization has become a focal point in algorithm quantization and accelerator design.
Previous studies on Transformer and Mamba quantization have typically selected target datasets based on specific application requirements or broad task categories.
In Transformer quantization, many approaches validate their methods using either a single dataset or task-specific dataset clusters.
For example, FQ-ViT employs COCO and ImageNet datasets to validate its 4-bit quantization approach for computer vision tasks, while Q-ViT uses only the ImageNet benchmark to evaluate fully quantized transformers for image classification~\cite{lin2021fq,li2022q}. 
Liu et al. evaluate their post-training mixed-precision quantization framework for ViT through a comparative study involving multiple datasets, including COCO, ImageNet, CIFAR-10, and CIFAR-100~\cite{liu2021post}.
In contrast, current quantization efforts for Mamba, represented by Quamba and MambaQuant, have been primarily limited to LLM scenarios, with evaluations conducted only on standard NLP benchmark datasets.

To our knowledge, only one hardware accelerator for Mamba has been published recently (MARCA~\cite{li2024marca}), and one will appear in DATE 2025~\cite{wei2025lightmamba}. 
Unlike our focus on resource-constrained edge applications, MARCA~\cite{li2024marca} is a reconfigurable accelerator that targets LLMs.
It accelerates element-wise operations that become increasingly dominant as sequence length grows, making it a critical component for LLMs, though it is less relevant for lightweight models.
Furthermore, its performance and area were evaluated through a simulator and synthesis tool, while we present a complete FPGA prototype and area evaluation using GF 22 nm technology. 
The second study, which is yet to be presented~\cite{wei2025lightmamba}, presents an FPGA-based accelerator with a quantization scheme that leverages a power-of-two scale for efficient re-quantization via shifting. 
Like MARCA, it targets LLMs and is evaluated using a large LLM model with 2.8B parameters, which is unsuitable for edge applications. 
In contrast, \emam is evaluated on three applications that can run in resource-constrained edge environments.
Furthermore, we propose a data-driven hybrid quantization approach: a scale-aware quantization for the SSM layer coupled with general 8-bit quantization for the other layers, aiming to achieve a more balanced quantization strategy. 
Unlike the existing two studies, we compare \emam against CNN, ViT, and na\"ive Mamba implementations and present hardware measurements on GF~22~nm. 
Finally, we emphasize that our framework is versatile and can be applied across a range of applications and other SSM-based machine learning architectures, such as S4~\cite{gu_Efficiently_2022}, S5~\cite{smith2022simplified}, and H3~\cite{fu2022hungry}.

\section{\emam Design framework Overview}\label{sec:overview}

\begin{figure*}[b] 
\centering
    \centering
    \vspace{-8pt}
    \includegraphics[width=0.99\linewidth]{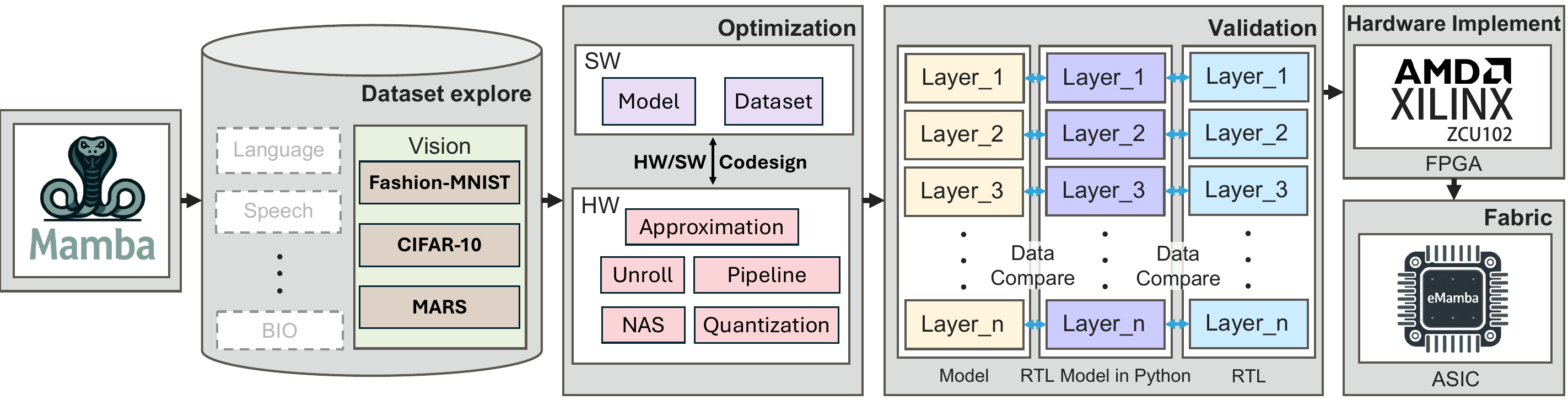}
    \vspace{-5pt}
    \caption{The proposed end-to-end design framework for \emam accelerator.}
    \label{fig:designflow}
\end{figure*}

Figure~\ref{fig:designflow} illustrates the proposed end-to-end design framework for \emam, targeting both FPGA and ASIC deployments. 
\emam can be demonstrated using various domains, including language, speech, and vision.
Among them, we focus on three representative vision datasets, Fashion-MNIST~\cite{fashionMNIST_2017}, CIFAR-10~\cite{cifar10_2009}, and MARS~\cite{an_MARS_2021}, due to their higher applicability to low-power embedded systems.

We propose a holistic optimization methodology to guide the accelerator design using the 
characteristics of the Mamba model and the target application using the following steps.

\begin{figure*}[t]
\centering
    \centering
    \includegraphics[width=0.99\linewidth]{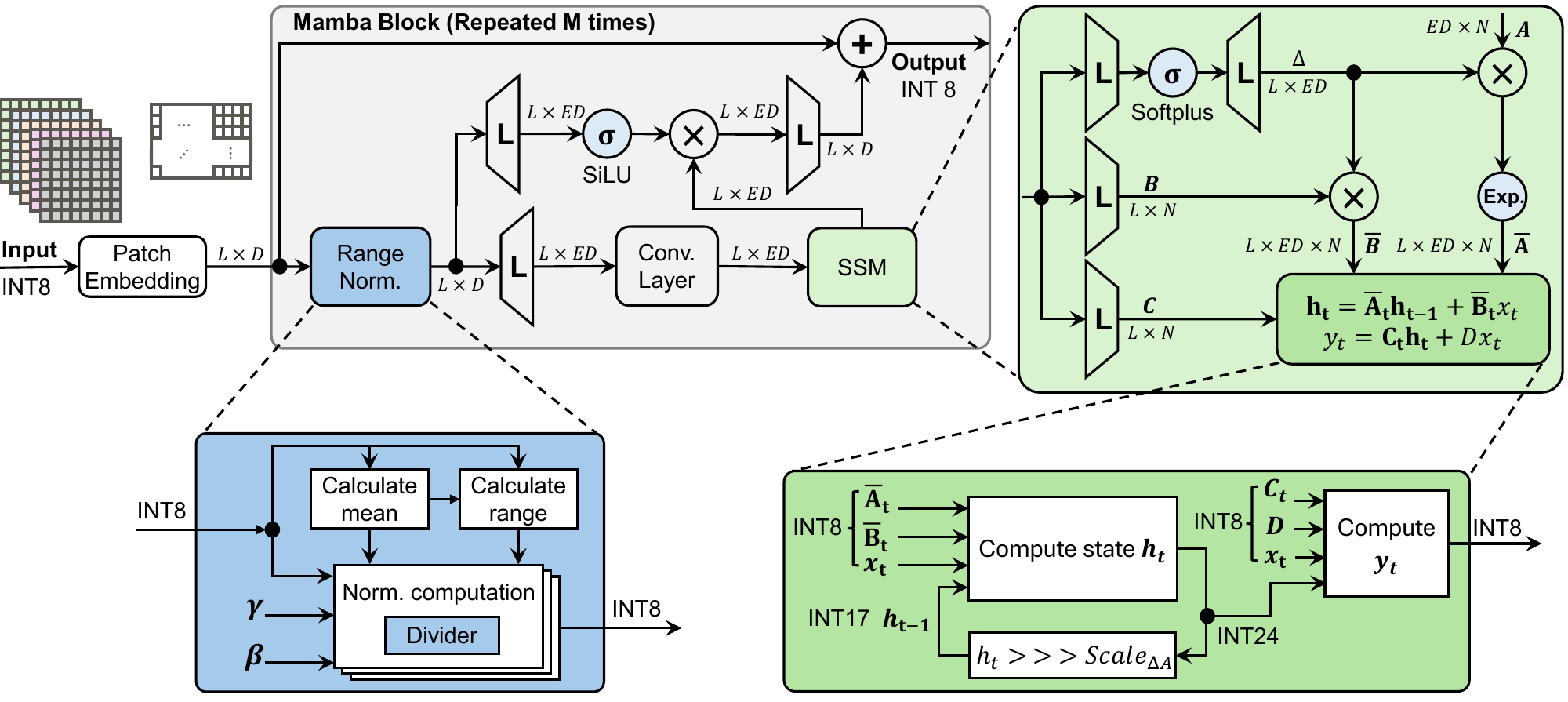}
    \vspace{-5pt}
    \caption{Overall architecture of the hardware-implemented Mamba block in \emam.}
    \vspace{-8pt}
    \label{fig:ssm_details}
\end{figure*}

\bh{1. HW optimization based on Mamba characteristics:} We first identify the computational bottlenecks and resource-intensive operations in the Mamba model for hardware acceleration, as highlighted in blue and green in Figure~\ref{fig:ssm_details}. 
Specifically, range normalization approximates layer normalization, which uses simpler arithmetic and light-weight parallelization (presented in Section~\ref{ssec:layer_norm}).
These approximations are configured by learnable parameters, such as $\gamma$ and $\beta$, as shown in Figure~\ref{fig:ssm_details}.
The SiLU activation is also replaced with a piecewise linear approximation to maintain both simplicity and nonlinearity (presented in Section~\ref{ssec:silu}).
We designed a custom pipeline for SSM computations, where recursive operations are unrolled and nonlinear functions such as Softplus and exponential are approximated with lightweight alternatives (detailed in Section~\ref{ssec:ssm}). 

\bh{2. Application-driven tuning on parameters and HW configurations:} The learnable parameters in approximation are fine-tuned based on the data distribution of the target application to optimize accuracy and robustness.
Then, layer-wise pipelining and NAS strategies are employed to guide architectural decisions (hyperparameters) that balance resource cost and accuracy for the given edge scenario (detailed in Sections~\ref{ssec:lw_pipe} and \ref{ssec:nas}).
Finally, we systematically quantize the floating-point operations through an application-driven analysis to determine the scale, bit-width, and clipping that minimize the computation overhead and hardware area under accuracy constraints (detailed in Section~\ref{ssec:quantization}).

After implementing the optimized architectural model, we employ a multi-stage validation procedure to accelerate the hardware design and verification process.
First, a high-level simulator is developed in Python to rapidly evaluate the functional correctness of the given architecture and ensure its accuracy on the target dataset.
Once the high-level model is validated, a cycle-accurate RTL implementation is developed and verified through RTL-level simulation.
After that, the RTL is synthesized and deployed on an FPGA for further functional testing and performance evaluation.
Finally, additional design steps and synthesis for the physical layout are conducted to prepare for the integration of custom silicon.
Detailed experimental results and validation procedures are presented in Section~\ref{sec:results}.

\section{HW Architecture Design and Optimization}\label{sec:design}

This section presents the details of the proposed Mamba architecture and application-driven optimizations outlined in the previous section.

\subsection{Hardware-friendly Range Normalization}\label{ssec:layer_norm}

Widely used layer normalization technique facilitates training stability and prevents issues like exploding or vanishing gradients~\cite{ba_Layer_2016}.
It normalizes the input samples ($x_i$) as:
\begin{equation}\label{layer_norm}
    \hat{y_i} = \gamma \cdot \frac{x_i - \mu}{\sqrt{\sigma^2 + \epsilon}} + \beta
\end{equation}
where $\mu$ and $\sigma^2$ represent the mean and variance of the input values, while $\epsilon$ is a small constant added for numerical stability.
The parameters $\gamma$ and $\beta$ are learnable parameters used for scaling and shifting.
The layer normalization involves square root and standard deviation computations, which are computationally expensive and inefficient for hardware targeting edge devices.

The computational efficiency of batch normalization in Equation~\ref{layer_norm} can be improved by approximating it using the \textit{range normalization}~\cite{banner_Scalable_2018} concept, which requires simpler operations.
However, employing the original concept sacrifices accuracy since it cannot capture input data dynamics accurately.
Therefore, \emam enhances range normalization with learnable parameters and simplifies hardware implementation while maintaining similar functionality as follows:
\begin{equation}\label{range_norm}
    \hat{y_i} = \gamma \cdot \frac{x_i - \mu}{range(x_i - \mu)} + \beta
\end{equation}
where $range(x) = max(x) - min(x)$, and $\gamma, \beta$ are learnable parameters.
The range normalization proposed in Equation~\ref{range_norm} requires comparators instead of square root and standard deviation computations, significantly improving computational efficiency while maintaining accuracy.

The division operation in Equation~\ref{range_norm} can be unrolled and parallelized across input $x$ to improve throughput.
However, division is a relatively costly operation in hardware, and extensive parallelism can introduce notable resource overhead.
We set up an evaluation scheme using FPGA implementations to quantify the trade-off between system-level performance gain and resource consumption.
During the hardware implementation, the Equation~\ref{range_norm} for each inference is decomposed into separate operations such as $mean$, $range$, and element-wise operations such as $devision$, $multiplication$, and $bit$-$shifting$.
The number of the total units for processing the normalization is parametric due to the different parallelization levels.
It allows us to identify an optimal design point that aligns with system-level constraints and application requirements.
For example, the parallelization level can be derived from its impact on this layer's latency and the system throughput.
The proposed \textit{approximation technique and optimized unrolling scheme} are essential for alleviating the normalization layer bottleneck in \emam and achieving an effective performance-area trade-off. 
Therefore, a detailed analysis and case study are presented in Section~\ref{ssec:rangeNorm}.

\subsection{Approximated Sigmoid Linear Unit Activation Design} \label{ssec:silu}

The Mamba architecture applies the SiLU activation in the secondary or residual path to the SSM, as shown in Figure~\ref{fig:ssm_details}.
The SiLU function combines sigmoid and linear units as:
\begin{equation}
    SiLU(x) = x\cdot \left( \frac{1}{1+e^{-x}}\right)
\end{equation}
\begin{figure}[b]
\centering
    \begin{minipage}{0.4\linewidth}
        \centering
        \includegraphics{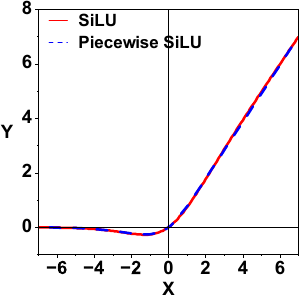}\\
        \small{(a) SiLU vs Piecewise SiLU}
        \label{fig:piecewise_exp}
    \end{minipage}
    \hspace{10mm}
    \begin{minipage}{0.4\linewidth}
        \centering
        \includegraphics{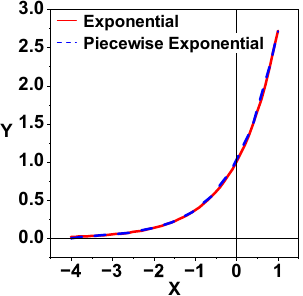}\\
        \small{(b) Exp. vs Piecewise Exp.}
        \label{fig:piecewise_silu}
    \end{minipage}
    \caption{Comparison of Piecewise exponential and Piecewise SiLU functions.}
    \label{fig:piecewise_functions}
\vspace{-5pt}
\end{figure}

This operation requires both exponentiation and division, making hardware implementation complex and resource-intensive.
Even though SiLU’s shape is similar to ReLU, replacing it with ReLU during training leads to a significant accuracy drop since \textit{SiLU is not a monotonically increasing function, unlike ReLU}.

\emam approximates SiLU as a piecewise linear function for inference.
Profiling the training data indicates a Gaussian distribution with a mean close to zero.
As Figure~\ref{fig:piecewise_functions}(a) illustrates, SiLU has the most substantial nonlinearity in the interval $[-7,7]$, where it decreases and then increases after reaching a minimum.
Hence, \emam uses more piecewise linear segments in this region and fewer segments away from the origin.
Furthermore, inputs less than negative seven are set to zero, while inputs greater than seven, where the linear part of SiLU dominates, are set to the input value.
We added discrete linear segments within the [-7,7] interval such that the error in any segment does not exceed $3\%$ compared to the original SiLU function, resulting in 17 linear segments.
This approximation does not cause a significant difference, as illustrated in Figure~\ref{fig:piecewise_functions}(a).

\subsection{SSM Layer with Hardware-Aware Approximation}\label{ssec:ssm}
The SSM layer is crucial for combining history and inputs to generate a new state and output.
Unlike typical neural network layers with intensive matrix-matrix multiplications, 
the SSM layer uses linear operations that are computationally more efficient.
It first uses the input $x_t$ to compute the $N$-dimensional state vector $\mathbf{h_t}$, defined in  Table~\ref{tab:notations}.
Then, it computes the output $y_t$ using the state vector and input as:
\begin{equation}\label{eq:ssm}
\begin{aligned}
    & \mathbf{h_t} = \mathbf{\bar{A}_{t}} \mathbf{h_{t-1}} + \mathbf{\bar{B}_{t}} x_t \\
    & y_t = \mathbf{C_t} \mathbf{h_t} + D x_t
\end{aligned}
\end{equation}

where $\mathbf{\bar{A}_{t}}$ is an $N$-dimensional diagonal matrix, 
while $\mathbf{\bar{B}_{t}}$ and $\mathbf{C_t}$ are $N$-dimensional vectors.
They are all functions of input $x_t$ since the step size $\mathbf{\Delta}$, and continuous time state-space vectors $\mathbf{B}$ and $\mathbf{C}$ are derived from the input $x_t$ using linear layers, as shown in Figure~\ref{fig:ssm_details}.
Specifically, $\mathbf{\bar{A}}$ and $\mathbf{\bar{B}}$ are calculated by multipying the $\mathbf{A}$ and $\mathbf{B}$ by the $\mathbf{\Delta}$, respectively.
This discretized SSM combined with the current input $x_t$ and past state $\mathbf{h_{t-1}}$ 
selectively encodes the input with history, replacing the transformer's attention mechanism by capturing selective history in the state vectors.

\begin{table}[b]
\caption{Key hyperparameters and variables used in this paper}
\label{tab:notations}
\renewcommand{\arraystretch}{0.9}
\begin{tabular}{@{}cl@{}}
\hline \hline
\textbf{Notations} & \textbf{Definition} \\ \midrule
$D$ & Model dimension or token size \\
$L$ & Sequence length \\
$E$ & Expansion factor ($ED$ is internal model dimension) \\
$N$ & State dimension \\
$P$ & Size of the patches \\
$M$ & Number of Mamba blocks \\
$\mu$, $\sigma^2$ & Mean and variance of the input to range normalization \\
$\epsilon$ & Stability constant \\
$\gamma$, $\beta$ & Learnable scale and shift in range normalization \\
$t$ & Time variable, goes from $0$ to $L$ \\
$x_t$, $y_t$ & Input and output from SSM at $t$ \\
$\mathbf{h_t}$ & State vector of size $N$ \\
$\mathbf{\bar{A}_{t}}$,$\mathbf{\bar{B}_{t}}$, $\mathbf{C_t}$ & Input dependent state-space matrices at $t$ \\
$\mathbf{\Delta}$ & Input dependent discretization parameter \\ 
\hline \hline
\end{tabular}
\end{table}

\emam implements the SSM operations in a pipelined manner with \textit{linear-time computation complexity}, processing one token (of size $ED$) at a time, where the input token of size $D$ is expanded to $ED$ through linear layers.
Since calculating $\mathbf{\bar{A}_{t}}$ and $\mathbf{\bar{B}_{t}}$ increases data size from $ED \times L$ to $ED \times N \times L$, processing entire sequences in parallel would consume substantial resources, where $N$ is the state dimension.
Hence, we use $ED \times N$ multipliers (in our example 40$\times$8) that iterate $L$ (in our example 16) times to calculate $\mathbf{\bar{A}_{t}}$ and $\mathbf{\bar{B}_{t}}$.
The first step involves element-wise multiplication: $x_t$ is multiplied by $\mathbf{\bar{B}_{t}}$, and the previous state vector $\mathbf{h_{t-1}}$ is multiplied by $\mathbf{\bar{A}_{t}}$.
The results are then added element-wise to produce $\mathbf{h_t}$.
Finally, the output is computed by summing up the dot product of $\mathbf{C_t}$ and $\mathbf{h_t}$ with the scalar $D x_t$.
While the SSM equation itself only requires multipliers and adders, 
preprocessing $\mathbf{\bar{A}}$ and $\mathbf{\bar{B}}$ involves non-linear functions such as softplus and exponential, as highlighted in Figure~\ref{fig:ssm_details}.
The implementation of these steps is detailed next.

\bh{Softplus to ReLU: }
The Softplus activation function is required to compute the discretization parameter $\mathbf{\Delta}$, as illustrated in Figure~\ref{fig:ssm_details} (in the highlighted SSM block).
Softplus is smooth and differentiable, but it is resource-intensive for hardware implementations. 
Since Softplus and ReLU are both monotonically increasing functions and Softplus closely resembles ReLU, 
\emam replaces Softplus with ReLU, a hardware-friendly activation function.
Experiments validate that switching to ReLU does not significantly reduce the accuracy of the Mamba model (presented in Section ~\ref{sec:results}).

\bh{Exponent design: }An exponential function is required to compute $\mathbf{\bar{A}}$ during the discretization process, as shown on the right-most side of Figure~\ref{fig:ssm_details}.
However, this function is computationally expensive for hardware implementation.
To address this, \emam adopts a piecewise linear approximation to reduce the complexity of the exponential function.
This technique, also used for approximating the SiLU function, involves profiling the training data to determine an effective input range for approximation.
Since most input values are less than one, values greater than one are treated as a constant, while inputs smaller than $-4$ are approximated as zero due to their negligible output.
The remaining range, from $-4$ to $1$, is approximated using 11 linear segments.
As shown in Figure~\ref{fig:piecewise_functions}(b), the piecewise approximation closely aligns with the original exponential function, resulting in minimal accuracy loss.

\subsection{Layer-Wise Pipelining Strategy in \emam} \label{ssec:lw_pipe}

\begin{figure*}[t]
\centering
    \centering
    \includegraphics[width=0.9\linewidth]{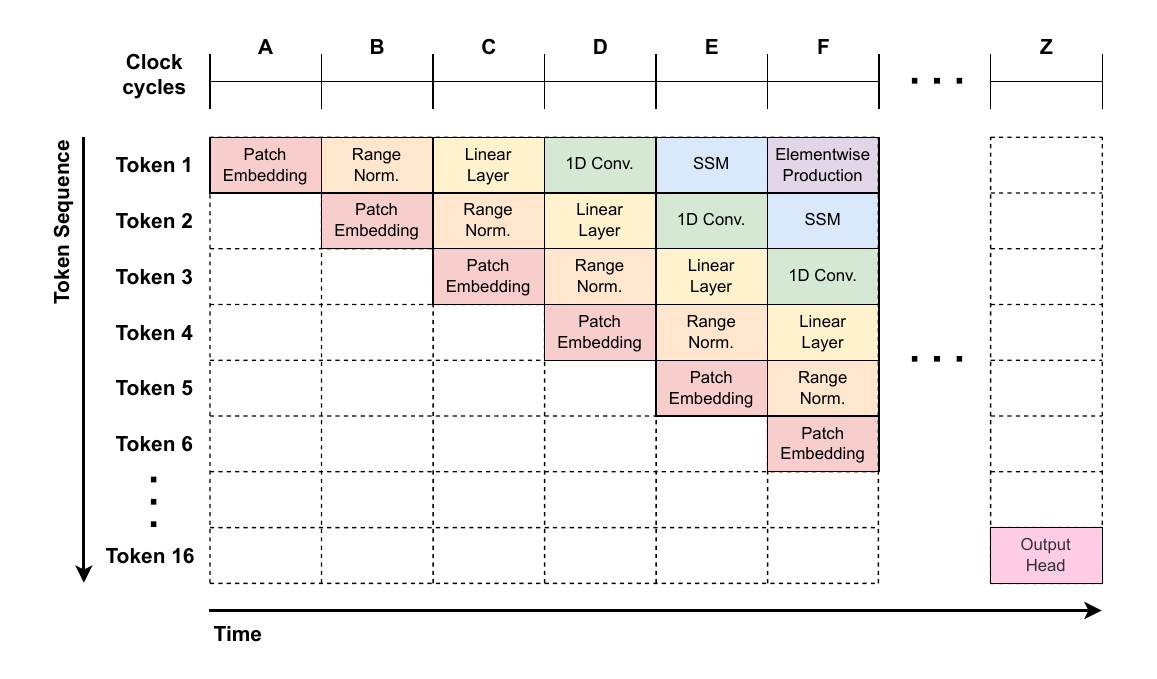}
    \vspace{5pt}
    \caption{Layer-wise pipelining schedule of \emam across abstract clock cycles. Each row corresponds to a token in the input sequence, and each column represents a synchronized pipeline stage labeled from Cycle~A to~Z. The colored blocks indicate key operations within each layer. To improve clarity, this figure presents a simplified view of the pipeline structure, omitting auxiliary or repeated operations.}
    \label{fig:pipeline}
    \vspace{-5pt}
\end{figure*}

\emam unrolls the Mamba computation and sets up a data flow architecture.
In contrast to the transformer’s attention mechanism, which requires all tokens to be available, SSM can process tokens sequentially, enabling linear token computation without needing the whole sequence of length $\mathbf{L}$.
To improve throughput while maintaining architectural simplicity, \emam employs a layer-wise pipelining approach across the computational blocks of Mamba.
As illustrated in Figure~\ref{fig:pipeline}, each layer operates as a pipeline stage, and tokens are propagated through these stages sequentially.
The execution time of each layer may be different since they may have different computational complexity and dimensionality.
Hence, \emam synchronizes the stages using a ready/valid handshake protocol.
Since each token is processed as a complete vector with a fixed dimension, partially processed or overlapping token inputs would violate the mathematical constraints of these operations.
As a result, \emam enforces strict token-by-token processing through each layer.
In this context, each layer begins processing a token as soon as it becomes available from the previous layer, without waiting for the entire sequence to complete.
This token-by-token flow allows continuous data movement through the network, improving throughput and enabling efficient hardware utilization.
Layers that complete early are stalled until the next stage is ready to accept the token, forming an uneven but coordinated pipeline flow.

For example, the input frames in the MARS dataset are partitioned into 16 non-overlapping patches, each treated as a token in the sequence.
These tokens are quantized and passed through the \emam one at a time.
While this approach takes more time to complete the first token than a fully asynchronous pipeline, it simplifies control logic and aligns well with hardware constraints such as timing closure.
Moreover, optimizations within SSM's recurrence and other computation-intensive layers reduce the critical path and improve pipeline balancing.

\subsection{Approximation-Aware NAS}\label{ssec:nas}

The \emam framework searches for the optimal neural architecture of Mamba by sweeping five key hyperparameters: ($D$, $E$, $P$, $N$, $M$), described in Table~\ref{tab:notations}.
As indicated in Figure~\ref{fig:ssm_details}, token size $D$ and its expansion factor $E$ are related to the internal dimension of the Mamba block, which can significantly impact the model's expressiveness and capacity.
$P$, the size of the patches, controls how the data is spatially partitioned into smaller segments.
It plays a key role in determining the sequence length $L$.
$N$ is state dimension that defines the shape of the parameters $\mathbf{\bar{A}}$, $\mathbf{\bar{B}}$ and $\mathbf{C}$.
Increasing the number of Mamba blocks $M$ may reduce the estimation error at the cost of the number of parameters.
We use range normalization and ReLU instead of layer normalization and softplus during training.
However, the piecewise linear approximation for the exponential and SiLU are applied only during the inference only because it changes the gradient of functions and hinders the proper training.

\begin{figure}[t]
\centering
    \centering
    \includegraphics[width=0.7\linewidth]{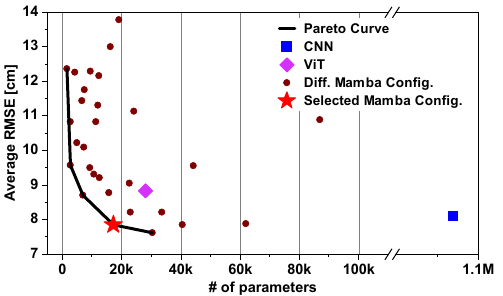}
    \caption{Results of approximation-aware NAS for the MARS dataset. The layer normalization and softplus are replaced with range normalization and ReLU, respectively.}
    \label{fig:hyperparam}
    \vspace{-5 pt}
\end{figure}

Figure~\ref{fig:hyperparam} shows the results of the NAS for the MARS dataset.
The x-axis is the number of parameters, while the y-axis is the average Root Mean Square Error (RMSE) compared to the ground truth.
We compare the results of various approximation-aware neural architectures of Mamba, represented by the brown dots, and CNN~\cite{an_MARS_2021} and ViT implementations.
There is a trade-off between the number of parameters and model accuracy, represented in terms of the average RMSE.
Among the Pareto front, we select a configuration, denoted by the red star, that has the smallest parameters while maintaining a similar accuracy obtained with the CNN.

\subsection{Quantization Optimization}\label{ssec:quantization}

\bh{Fully quantized \emam: }Integer operations significantly reduce computational complexity, save memory, and enhance overall processing performance.
Therefore, fully quantizing the model in inference is a promising method to save memory and computational resources for efficient deployment on edge devices~\cite{wu_Training_2018}.
\emam aims for an efficient accelerator using integer operations across the design instead of traditional floating-point operations.
Equation~\ref{eq:quantization} shows how a floating point variable $x$ is quantized into a fixed point integer:
\begin{equation} \label{eq:quantization}
    Q(x) = \text{clamp}\left(\left\lfloor \frac{x - Z}{S} \right\rceil, \text{min}_{\text{range}}, \text{max}_{\text{range}}\right)
\end{equation}
where \( Z \) is the zero-point, 
\( S  \) is the scale factor that determines the interval between adjacent quantized values.
Finally, minimum and maximum ranges \( \text{min}_{\text{range}} \) and \( \text{max}_{\text{range}} \) are determined by the bit-width \( b \) such that \( \text{min}_{\text{range}} = -2^{b-1} \) and \( \text{max}_{\text{range}} = 2^{b-1} - 1 \).

We profile the data distribution in the Mamba model to determine the quantization parameters, including the scale factor and zero point.
Observing that most activations and weights are concentrated around zero, we set the zero point to 0 and employed symmetric uniform quantization \cite{nagel_White_2021}.
\emam restricts the scale factors as powers of two, enabling the transformation of floating-point multiplications into integer multiplications with bit shifts, thus facilitating efficient hardware implementation.

Based on the data distribution in each layer, different scaling factors 
$S$ are applied using shift operations to effectively cover the majority of values.
Clipping is employed to handle the small number of outliers efficiently.
\emam applies the same number of bits for quantization to parameters of all layers to simplify the hardware architecture.
We examine how accuracy changes when the bit widths of weights, biases, and activations vary.
We check all combinations of 8-bit, 16-bit, and 32-bit and confirm that the bit-widths of weights and biases have negligible accuracy impact.
The bit-width of activation affects the model accuracy, but the accuracy drop is insignificant.
So, we use 8-bit integers for all parameters rather than designing complex schemes that quantize each parameter with a different bit-width.

\bh{Scale-aware quantized SSM layer: }
We implement the quantized SSM to reduce the complexity of computation.
However, applying fixed bit-width quantization to all intermediate hidden states in the SSM leads to significant accuracy degradation.
The SSM computation block uses a higher bit-width for those intermediate hidden states.
Compared to floating-point arithmetic, quantized integer multiplication must explicitly account for the respective scaling factors of operands.
When computing \(h_t\), the previous state \(h_{t-1}\) is multiplied by the parameter \(\bar{A}_{t}\), and their scaling factors are also multiplied accordingly.
Due to the SSM's recurrent nature, this operation is repeated over the entire sequence, resulting in an explosion of bit width.
Therefore, \(h_t\) requires proper tuning of bit-width and re-quantization to prevent overflow and maintain numerical stability.

Re-quantization helps to maintain numerical stability, but it loses the information of \(h_t\).
This causes a problem because it is used to compute \(y_t\) of the current token.
We use \(h_t\) to compute \(y_t\) before the re-quantization and store the re-quantized \(h_t\) as \(h_{t-1}\) to calculate new \(h_t\) in the subsequent step.
Specifically, \(h_t\) is divided by the scaling factor of \(\bar{A}_t\) before storing it, thereby preventing bit-width accumulation across time steps.

For example, in the MARS dataset, \(\bar{A}_t\) has a fixed scale of \(2^{-7}\); hence, the INT24 result is right-shifted by 7 bits and stored as INT17.
\(h_t\) with 24 bits is used to compute \(y_t\), and \(y_t\) is quantized to INT8.
All other inputs, outputs, and parameters of the SSM block, except for \(h_t\), are constrained to INT8 to reduce area cost.

\section{Experimental Results}\label{sec:results}

\subsection{Experimental Setup}

\bh{Datasets: }\label{ssec:dataset}
The proposed \emam architecture is evaluated on four datasets:

\begin{itemize}[leftmargin=*]
   
\item \textit{Fashion-MNIST~\cite{fashionMNIST_2017}}:
This drop-in replacement for MNIST consists of 70,000 grayscale 28×28 images of fashion items, evenly split into training and test sets.
Accuracy is measured by the proportion of correctly classified images. 

\item \textit{CIFAR-10~\cite{cifar10_2009}}
is a commonly used labeled image dataset that contains 60,000 32×32 color images across 10 object classes, 50,000 of which are for training and 10,000 for testing.
Model performance is evaluated using classification accuracy.
It is used instead of more complex models since we target low-power edge devices.

\item \textit{MARS~\cite{an_MARS_2021}} is an open-source dataset designed for mmWave-based human pose estimation.
It contains 40,083 frames of 3D pointcloud data capturing human movement, with 80\% used for training/validation (further split 8:2) and 20\% for testing.
Each inference produces 57 values corresponding to the 3D coordinates of 19 human joints.
Accuracy is evaluated using mean absolute error (MAE) and RMSE.

\item \textit{WikiText2~\cite{wikitext2_2016}} is a word-level language modeling dataset extracted from high-quality Wikipedia articles.
It consists of approximately 2 million tokens spanning 720 articles, divided into 600 for training, 60 for validation, and 60 for testing.
Model performance is evaluated using perplexity.
\end{itemize}

\begin{table}[b]
\caption{Hyperparameter configurations of ViT and \emam for each dataset}
\label{tab:hyperparameters}
\begin{tabular}{c|ccccc|ccccc}
\hline \hline
\multirow{3}{*}{Hyperparam.} & \multicolumn{5}{c|}{ViT}                       & \multicolumn{5}{c}{\emam}               \\ \cline{2-11} 
                             & Model & MLP  & Patch & Transformer & Number of & D  & E & P & M & N  \\
                             & size  & size & size  & Layers      & heads     & \multicolumn{5}{c}{(Defined in Table~\ref{tab:notations})} \\ \hline
Fashion-MNIST                & 128   & 128  & 6     & 8           & 6         & 24  & 2 & 2 & 4 & 16  \\
CIFAR-10                     & 64    & 512  & 8     & 6           & 8         & 64  & 2 & 4 & 4 & 32  \\
MARS                         & 32    & 128  & 2     & 2           & 4         & 20  & 2 & 2 & 2 & 8   \\
WikiText2                    & 100   & 200  & -     & 2           & 4         & 100 & 2 & - & 2 & 128 \\\hline \hline
\end{tabular}
\end{table}

\bh{Baseline for the comparisons:}
The proposed \emam architecture is compared against two baseline models: a ViT model and a reference Mamba model without any proposed approximations (\textit{Ref. Mamba}) in three configurations tailored to three datasets.
The first group is tailored to Fashion-MNIST~\cite{learnable_vit_2023}, the second is tailored to CIFAR-10~\cite{understanding_vit_2023}, and the third is tailored to MARS with comparable state-of-the-art accuracy~\cite{an_MARS_2021}.
Table~\ref{tab:hyperparameters} summarizes the hyperparameter configurations for ViT and \emam. The \textit{Ref. Mamba} uses the same hyperparameters as \emam.
We configure the models by the result of NAS, as described in Section~\ref{ssec:nas}. 
All the comparisons are performed using floating point (FP32) and 8-bit quantized (INT8) precision, where all weights, biases, and activations of post-trained models are quantized using the Brevitas framework~\cite{brevitas}.
In addition, for the WikiText2 dataset, we include RNN\cite{rnn_2010} and LSTM~\cite{lstm_2014} models as representative baselines to evaluate the effectiveness of \emam on long-sequence natural language tasks.

\subsection{Accuracy and Model Size Evaluation}

\begin{table}[t]
\caption{Comparison of ViT and \emam in terms of accuracy and model size under FP32 and INT8 quantization on Fashion-MNIST, CIFAR-10, and MARS datasets}
\label{tab:acc_comp_3dataset}
\begin{tabular}{c|c|cc|cc|cc}
\hline \hline
\multirow{2}{*}{}     & \multirow{2}{*}{Model} & \multicolumn{2}{c|}{Fashion-MNIST}     & \multicolumn{2}{c|}{CIFAR-10}   & \multicolumn{2}{c}{MARS}  \\ \cline{3-8} 
    & & \multicolumn{1}{c|}{\# of bytes} & Acc. {[}\%{]} & \multicolumn{1}{c|}{\# of bytes} & Acc. {[}\%{]} & \multicolumn{1}{c|}{\# of bytes} & RMSE {[}cm{]} \\ \hline
\multirow{2}{*}{FP32} & ViT & \multicolumn{1}{c|}{3.04M}    & 87.6  & \multicolumn{1}{c|}{1.95M}    & 77.5  & \multicolumn{1}{c|}{110K} & 7.38  \\
    & \emam & \multicolumn{1}{c|}{157K} & 90.2  & \multicolumn{1}{c|}{988K} & 78.3  & \multicolumn{1}{c|}{67.3K}    & 7.85  \\ \hline
\multirow{2}{*}{INT8} & ViT & \multicolumn{1}{c|}{778K} & 86.4  & \multicolumn{1}{c|}{499K} & 75.6  & \multicolumn{1}{c|}{27.4K}    & 9.05  \\
        & \emam & \multicolumn{1}{c|}{39.1K}    & 86.5  & \multicolumn{1}{c|}{247K} & 72.6  & \multicolumn{1}{c|}{16.8K}    & 8.83   \\ \hline \hline
\end{tabular}
\end{table}

\bh{Comparisons against the VIT models:} 
\emam is evaluated against ViT models across three datasets under FP32 and INT8 quantization, as shown in Table~\ref{tab:acc_comp_3dataset}.
Although no approximation techniques are applied to the ViT models, \emam outperforms ViT on Fashion-MNIST (90.2\% vs. 87.6\%) while using 19.9$\times$ fewer parameters (3.04M vs. 157K) with FP32, as indicated in Table~\ref{tab:acc_comp_3dataset}.
On CIFAR-10 and MARS, \emam achieves comparable performance (78.3\% vs. 77.5\% and 7.85 cm vs. 7.38 cm), while reducing model size by 2.02$\times$ and 1.63$\times$, respectively.
Under INT8 quantization, both models exhibit slight accuracy degradation, as shown in Table~\ref{tab:acc_comp_3dataset}. On Fashion-MNIST, \emam continues to outperform ViT (86.5\% vs. 86.4\%).
On CIFAR-10, \emam shows slightly more degradation (72.6\% vs. 75.6\%), which can be attributed to its sensitivity to channel-wise statistical variation under quantization, as mentioned.
For the MARS dataset, \emam performs better (8.83 cm vs. 9.05 cm) again.
Both quantized models are proportionally scaled down from their FP32 counterparts, maintaining the same relative reduction ratio between ViT and \emam.

\bh{\emam with complex task:}

\begin{figure}[t]
\centering
\includegraphics[width=0.7\linewidth]{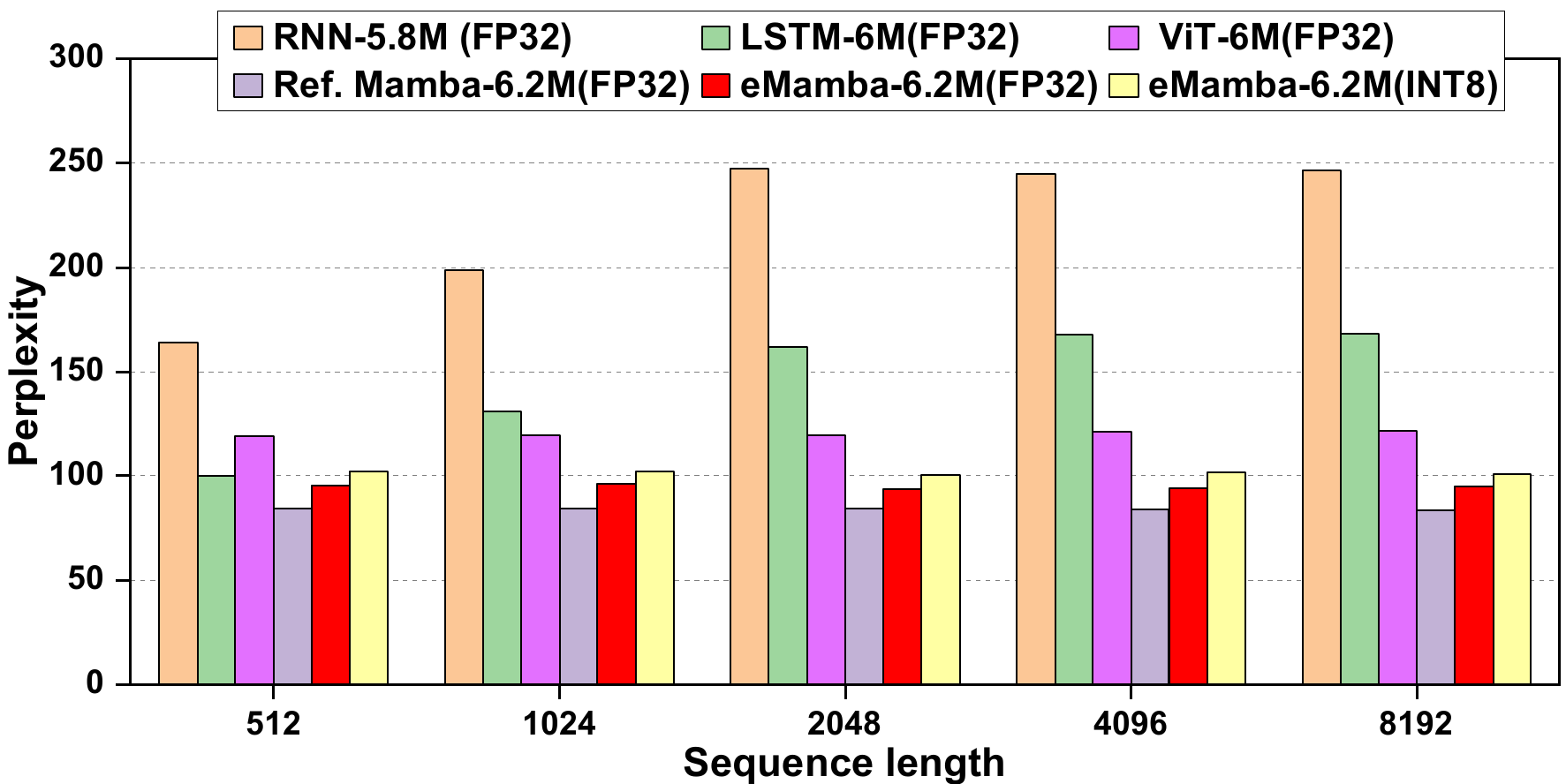}
\caption{Perplexity changes according to sequence length in the WikiText2 dataset.}
\label{fig:wikitext2}
\end{figure}

In addition to small vision tasks such as CIFAR-10, Fashion MNIST, and pose detection, we have evaluated \emam on the WikiText2 dataset to show its broader applicability on complex tasks.
Figure~\ref{fig:wikitext2} shows the changes in perplexity according to sequence lengths for several models of comparable size on WikiText2. To ensure a fair comparison, all baseline models, including RNN, LSTM, ViT, \textit{Ref. Mamba}, and \emam, are configured to have a similar model size of approximately 6M parameters, matching that of RNN and LSTM.
When the sequence length increases from 512 to 8,192, \emam maintains a consistently low perplexity and even slightly outperforms a ViT, as shown in Figure~\ref{fig:wikitext2}.

For example, at a sequence length of 512, \emam in FP32 achieves a perplexity of 95.61, which, while relatively high compared to large language models, is attributed to the limited model size.
Across all sequence lengths, \emam demonstrates greater stability than the RNN~\cite{rnn_2010} and LSTM~\cite{lstm_2014} models, whose perplexities increase sharply as the sequence length increases.
RNN increases from 163.91 at 512 to 246.45 at 8192, and LSTM from 99.99 to 168.42. In contrast, the perplexity of ViT, \textit{Ref. Mamba}, and \emam with FP32 remain nearly flat, from 119.4 to 121.8, 83.55 to 84.25, and 95.61 to 94.69, respectively.
Finally, the INT8 quantized version of \emam achieves perplexities nearly identical to its FP32 model, such as 100.8 at a sequence length of 512 and 102.4 at 8192, highlighting its suitability for deployment in resource-constrained devices.
\emam demonstrates considerable effectiveness in handling long sequences compared to other architectures.
These results confirm that \emam not only excels in small-scale vision tasks but also generalizes effectively to long-sequence NLP tasks, making it a versatile backbone for diverse deep neural network applications.
Nevertheless, considering our target deployment scenario for resource-constrained edge applications, we select the MARS dataset to evaluate the effectiveness of \emam in terms of accuracy, frame latency, throughput, and area.

\begin{table}[b]
\tabcolsep=0.17cm
\centering
    \caption{Accuracy and model size comparison among the baseline CNN, ViT, Mamba without approximation, and \emam on MARS dataset.}
    \label{tab:accuracy_results}
    \begin{tabular}{@{}cc|cccc|ccc@{}}
    \hline \hline
    &  & \multicolumn{4}{c|}{FP32} & \multicolumn{3}{c}{INT8 } \\ \cline{3-9} 
    &  & CNN~\cite{an_MARS_2021} & \hspace{-2mm}ViT~\cite{ViT_iclr_2021} & \hspace{-2mm}Ref. Mamba\hspace{-2mm} & \emam & CNN & ViT & \emam \\ \hline
    \multicolumn{2}{c|}{\# of bytes} & 4.14M & 110K & 67.3K & 67.3K & 1.03M & 27.4K & 16.8K \\ \hline
   \multirow{2}{*}{\begin{tabular}[c]{@{}c@{}}X\\ {[}cm{]}\end{tabular}}
   & MAE   & 6.97  & 5.76 & 6.38   & 6.64  & 7.74  & 7.41 & 7.21  \\
   & RMSE  & 9.85 & 8.63 & 9.15  & 9.46  & 10.7 & 10.6  & 9.97  \\ \hline
   \multirow{2}{*}{\begin{tabular}[c]{@{}c@{}}Y\\ {[}cm{]}\end{tabular}} 
   & MAE   & 3.83 & 3.51 & 3.62  & 3.74  & 4.78 & 7.03 & 5.40  \\
   & RMSE  & 5.49 & 4.94 & 5.04  & 5.18  & 6.58 & 6.86 & 6.94  \\ \hline
   \multirow{2}{*}{\begin{tabular}[c]{@{}c@{}}Z\\ {[}cm{]}\end{tabular}}
   & MAE   & 6.99 & 6.16 & 6.47  & 6.61  & 7.49 & 7.03 & 7.29 \\  
   & RMSE  & 9.38 & 8.56 & 8.81  & 8.92  & 9.97 & 9.67 & 9.58  \\ \hline
   \multirow{2}{*}{\begin{tabular}[c]{@{}c@{}}\textbf{Average}\\ {[}cm{]}\end{tabular}}& \textbf{MAE}   & \textbf{5.93} & \textbf{5.14} & \textbf{5.49}  & \textbf{5.66}  & \textbf{6.67} & \textbf{6.59} & \textbf{6.63}  \\
   & \textbf{RMSE}  & \textbf{8.24} & \textbf{7.38} & \textbf{7.67}  & \textbf{7.85}  & \textbf{9.08} & \textbf{9.05} & \textbf{8.83}  \\ \hline \hline
   \end{tabular}
\end{table}

\bh{More detailed analysis on the MARS dataset:}
Table~\ref{tab:accuracy_results} presents more detailed comparisons of \emam against the CNN and ViT~\cite{understanding_vit_2023, learnable_vit_2023} models, where ViT and \emam are configured as shown in Table~\ref{tab:hyperparameters}.
\emam outperforms the CNN model (5.66 cm MAE vs. 5.93 cm MAE and) using 63$\times$ smaller model size (67.3 KB vs. 4.14 MB).
Moreover, the quantized \emam achieves better accuracy than the quantized CNN (6.63 cm MAE vs. 6.67 cm MAE and 8.83 cm RMSE vs. 9.08 cm RMSE), while reducing the model size to 16.8 KB, again 63$\times$ smaller than the quantized CNN model.
Convolution is more parallelizable and data-efficient than the operation in a sequential model. However, model compactness, competitive accuracy, and linear-time complexity make \emam a strong candidate for resource-constrained edge applications.

The results of ViT and \emam remain consistent with earlier comparisons.
Additionally, Table~\ref{tab:accuracy_results} shows that the MAE result of \emam is slightly worse than ViT in FP32 and INT8 cases, using 1.63$\times$ less model size.
However, ‌\emam maintains linear computational complexity (O(n)) against ViT’s quadratic scaling (O(n²))‌, a decisive advantage for real-time edge computing~\cite{gu_Mamba_2023}, which directly impacts the hardware performance compared with \emam. Based on this accuracy, model size, and hardware efficiency analysis of these models, we further compare \emam hardware against CNN and ViT accelerators in terms of throughput, latency, and resource cost, as discussed in Section~\ref{ssec:perf}.

\subsection{Design Space Exploration of Resource-Aware Range Normalization}\label{ssec:rangeNorm}
Before analyzing the resource requirements of \emam in detail, we first explore the design space of the range normalization layer.
Unlike other layers, range normalization is primarily an element-wise operation, making it well-suited for various parallelization strategies.
We deploy the range normalization layer on the AMD Zynq UltraScale+ MPSoC ZCU102 board \cite{zcu102board} using AMD Vivado 2024.1, where the arithmetic logic of it is synthesized into DSP48E2 blocks. 
These blocks efficiently handle arithmetic operations except for division. 
Thus, each compute unit in the range normalization layer requires one DSP block and a dedicated divider.
We implement it using 1, 2, 4, 5, 10, and 20 compute units to explore the trade-off between latency and resource utilization.
These configurations were chosen due to the token dimension (20), allowing for a uniform workload distribution across units.
One compute unit requires 25~cycles per element: 2~cycles to compute the mean and range, and 23~cycles to perform division, multiplication, addition, and bit-shifting.
The number of compute units is important because it affects the frame latency.

\begin{figure}[b!]
\centering
\includegraphics[width=0.65\linewidth]{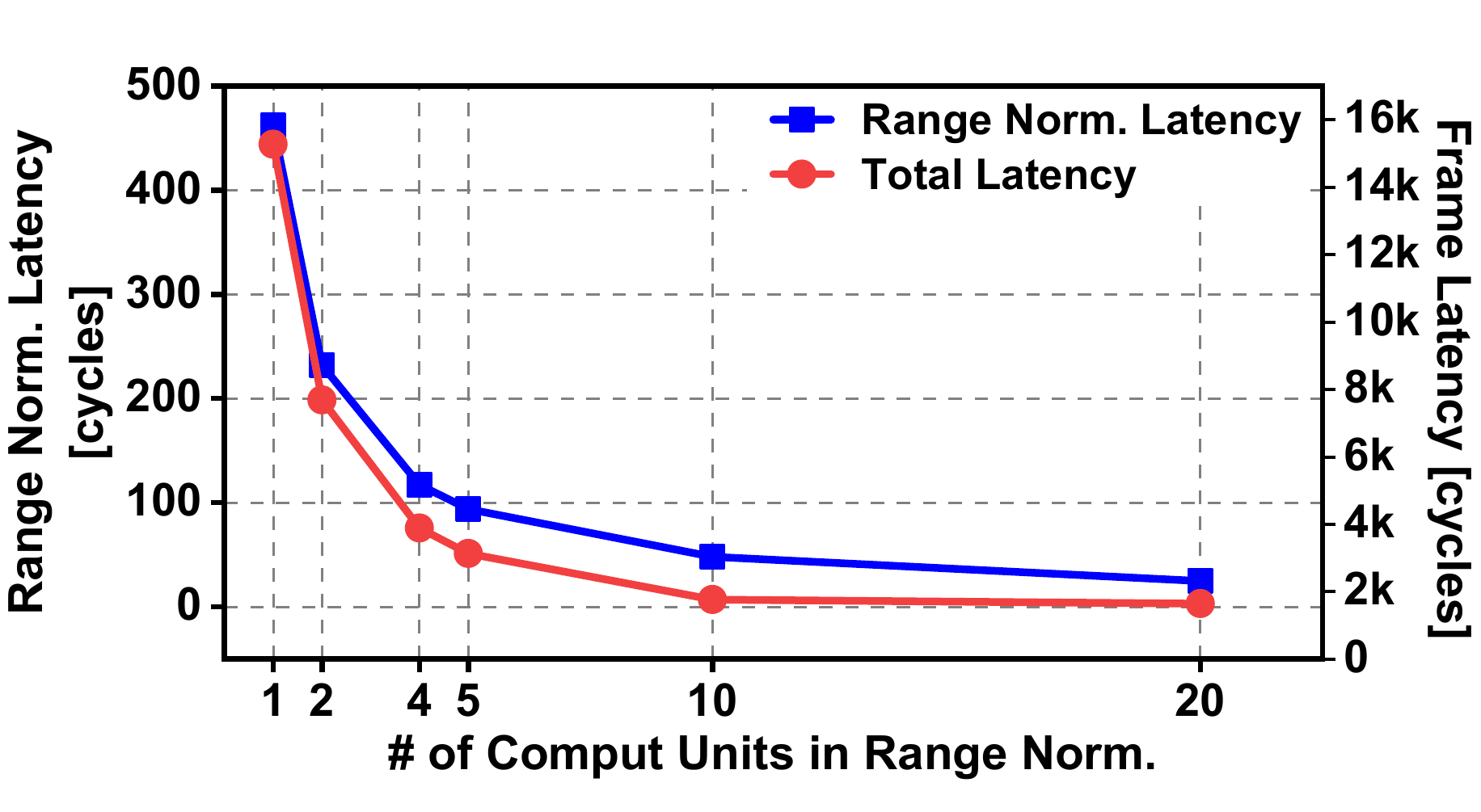}
\vspace{-3mm}
\caption{Range normalization and frame latency with respect to the number of compute units.
Latency decreases with increased parallelism, with diminishing returns observed beyond 10 units.}
\label{fig:rangeNorm_cycles}
\end{figure}

Figure~\ref{fig:rangeNorm_cycles} shows the number of cycles to calculate one element in range normalization and frame latency according to the number of compute units in range normalization.
Increasing the number of compute units from 1 to 10 yields an almost linear reduction in both range normalization latency and frame latency.
This trend highlights range normalization as the dominant bottleneck within this configuration range.
Beyond 10 units, however, further latency reduction becomes marginal.
This is due to other stages in the pipeline, requiring approximately 40~cycles per token, while the final projection layer, targeting the output dimension of the MARS dataset, takes 58~cycles. 
Furthermore, as range normalization is positioned at the beginning of the Mamba block, its latency determines the pipeline stage latency and limits the overall pipeline efficiency.
Furthermore, the resource utilization increases linearly with the number of units, as shown in Table~\ref{tab:rangenorm_hwutilization}, since each unit operates independently.

Based on these observations, configurations with fewer than five units introduce substantial latency penalties that outweigh resource savings.
Hence, our \emam implementation for the MARS dataset focuses on the 10- and 20-unit configurations.
While increasing the number of units requires more hardware resources, the overhead remains minimal relative to the total available resources.
Specifically, the additional resource usage accounts for only 0.66\% of LUTs, 0.09\% of FFs, and 0.79\% of DSP slices on the ZCU102 platform, making it a worthwhile trade-off.
Given the 1.92$\times$ reduction in range normalization latency and a 1.08$\times$ improvement in frame latency, the 20-unit configuration is selected to fully parallelize range normalization with minimal hardware cost.

\begin{table}[t]
\centering
\captionof{table}{The hardware resource utilization for different numbers of compute units in range normalization, expressed as percentages of ZCU102 platform capacity}
\label{tab:rangenorm_hwutilization}
\vspace{-2mm}
    \begin{tabular}{c|cccccc}
    \hline \hline
    \# of compute units in range norm. & 1    & 2    & 4    & 5    & 10   & 20   \\ \hline
    LUT [\%]                                   & 0.07 & 0.13 & 0.26 & 0.33 & 0.66 & 1.32 \\
    FF [\%]                                    & 0.01 & 0.02 & 0.04 & 0.04 & 0.09 & 0.18 \\
    DSP [\%]                                   & 0.08 & 0.16 & 0.32 & 0.40 & 0.79 & 1.59 \\
    \hline \hline
    \end{tabular}
\vspace{-10pt}
\end{table}

\subsection{Performance Evaluation on FPGA} \label{ssec:perf}

This section evaluates \emam on Fashion-MNIST, CIFAR-10, and MARS datasets and compares it against the CNN and ViT baselines.
We focus on hardware evaluation using the MARS dataset, a representative edge application, since
the overall hardware architecture and computation flow remain essentially unchanged across tasks and repetitive hardware implementations are redundant.

We implement two configurations of \emam: the reconfigurable \emam, which can change the value of weights and biases at runtime, while the embedded \emam, which has fixed weights and biases.
The reconfigurable version provides more flexibility than the embedded version.
The embedded version (using fixed weights and biases) may be preferred when the implementation is tailored to a given application, and reprogramming the FPGA is practical when the dataset changes.
The proposed techniques facilitate adapting \emam models for other applications and use cases.

In addition, we use the FINN framework~\cite{blott2018finn,finn} to implement the quantized CNN accelerator
and a ViT accelerator based on I-ViT~\cite{li2023vit} on the same FPGA, applying the same quantization and hardware constraints used in \emam.
We further implement a na\"ive version of the Mamba model (\textit{Na\"ive Mamba}) using simplified components from I-ViT and MARCA~\cite{marca_2024} to serve as a hardware baseline.
Specifically, we adopt the I-layer normalization, shift-based exponential approximations from I-ViT, and the piecewise SiLU optimization from MARCA.
The softplus function is implemented using a 22-segment linear piecewise approximation.
The remaining layers follow the same configurations used in \emam, as components such as linear, convolution and element-wise operations can be efficiently implemented with integer arithmetic.
In contrast, exact implementations of functions like normalization and SiLU are computationally expensive and impractical to realize with integers, which motivates the use of approximations for those operations.
All designs have a 100 MHz operating clock frequency.
The CNN throughput configuration in the FINN compiler is incrementally adjusted until the post-implementation LUT utilization becomes comparable to reconfigurable \emam.

Table~\ref{tab:performance_results} lists two performance metrics: frame latency and throughput.
Frame latency is the number of cycles from the receipt of the pointcloud to get a single inference output.
Throughput quantifies the number of inferences the accelerator can make per second in a pipelined manner.
Both versions of \emam take 1,643 cycles to process a frame, leading to 16.43 $\mu$s inference time at 100 MHz clock frequency.
This speed is 5.62$\times$ faster than the CNN and 4.95$\times$ faster than the ViT accelerators, and also achieves a 6.22$\times$ reduction in frame latency compared to the na\"ive Mamba baseline.
Similarly, \emam processes almost 263~Mb/s, which is 9.95$\times$ and 2.22$\times$ higher than the CNN and ViT accelerators, respectively, and 5.32$\times$ higher than na\"ive Mamba.
This significant performance enhancement is largely due to the low computational complexity and decreased model size of \emam.
This improvement over na\"ive Mamba is primarily due to replacing I-layer normalization with range normalization, which alleviates the bottleneck in the pipelined architecture.
Detailed analysis is presented in Section~\ref{ssec:ablation}.

\begin{table}[t]
\setlength{\tabcolsep}{4pt}
\renewcommand{\arraystretch}{1.2}
\centering
\caption{Performance comparison of 8-bit quantized CNN, ViT, Na\"ive Mamba, and \emam}
\label{tab:performance_results}
\begin{tabular}{@{}l|cccc|ccc@{}}
\hline \hline
\multirow{3}{*}{} 
& \multirow{3}{*}{\textbf{CNN}} 
& \multirow{3}{*}{\textbf{ViT}} 
& \multirow{3}{*}{\shortstack{\textbf{Na\"ive} \\ \textbf{Mamba}}} 
& \multirow{3}{*}{\textbf{\emam}} 
& \multicolumn{3}{c}{\textbf{Gain}} \\
\cline{6-8}
& & & & & \textbf{CNN} & \textbf{ViT} & \shortstack{\textbf{Na\"ive} \\ \textbf{Mamba}} \\
\hline
\textbf{Frame Latency {[}cycles{]}} & 9,235 & 8,130 & 10,220 & 1,643 & 5.62$\times$ & 4.95$\times$ & 6.22$\times$ \\
\textbf{Throughput {[}Mb/s{]}}      & 26.4  & 118   & 49.4   & 263   & 9.95$\times$ & 2.22$\times$ & 5.32$\times$ \\
\hline \hline
\end{tabular}
\end{table}

\subsection{Resource Utilization}

\begin{figure}[b]
\centering
    \centering
    \includegraphics{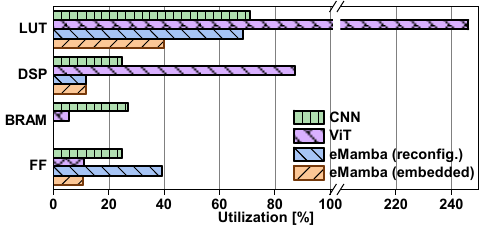}
    \caption{Resource usage comparison of 8-bit quantized CNN, ViT and \emam.}
    \label{fig:utilization_resource}
\end{figure}

As described in Section~\ref{ssec:perf}, we configure the baseline CNN accelerator to match the LUT usage of the reconfigurable \emam.
We also implement a ViT accelerator under the same quantization and hardware constraints.
The reconfigurable and embedded \emam consume 68.5\% and 40.0\% of LUTs, respectively, while the CNN accelerator uses 71.0\%, as shown in Figure~\ref{fig:utilization_resource}.
In contrast, the ViT accelerator consumes 246\% of LUTs, making it impossible for deployment on the ZCU102 platform without significant architectural modification.
The reconfigurable \emam stores weights in FFs, whereas the embedded \emam stores them in LUTs.
Due to the use of FFs, the reconfigurable version requires additional control logic for weight access and update, which is implemented using LUTs.
As a result, it consumes more LUTs than the embedded \emam.

Both \emam implementations use only 11.8\% of DSPs, while the CNN implementation uses 24.8\% since our carefully optimized pipeline effectively uses LUTs and FFs.
In contrast, the ViT accelerator consumes 87.3\% of DSPs despite applying pipelining technique similar to \emam due to its higher computational complexity.
In terms of memory blocks, both \emam implementations do not use any BRAMs, whereas the CNN and ViT accelerator consumes 247 and 50 BRAM blocks, respectively.
Regarding FFs, the reconfigurable and embedded \emam use 39.1\% and 10.8\% of the FFs, respectively, while the CNN and ViT accelerators use 24.9\% and 11.1\%, respectively.
In comparison, reconfigurable \emam uses FFs for both pipeline implementation and weight storage, leading to higher FF usage compared to the CNN accelerator, which relies on BRAM.
These comparisons demonstrate that both \emam implementations are more memory-efficient and scalable for edge applications than the CNN and ViT accelerator.

\subsection{Hardware Implementation Comparison on GF 22~nm Technology}

\begin{figure}[b]
\centering
\includegraphics[width=0.4\columnwidth]{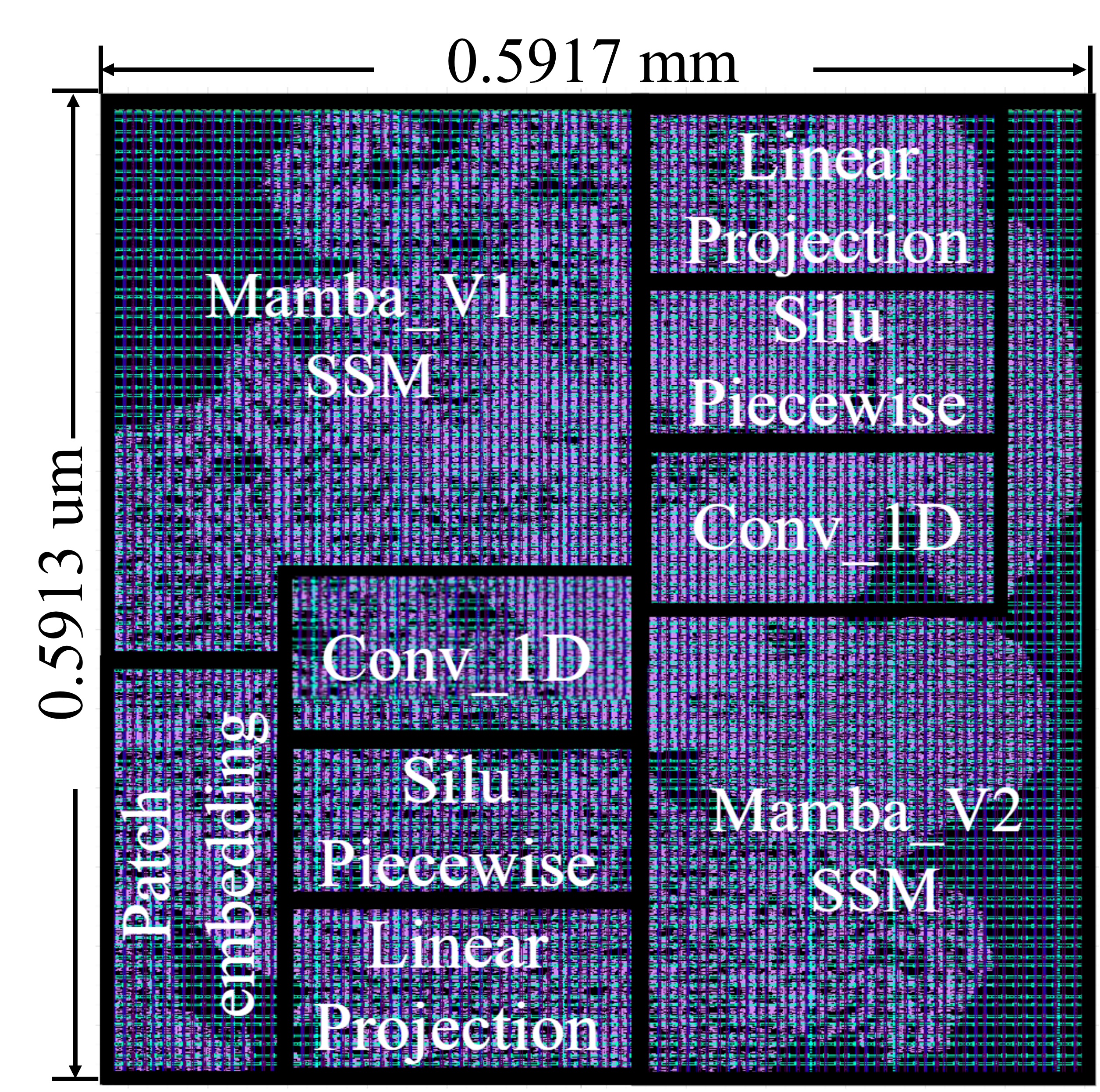}
\caption{Layout of the designed \emam.}
\vspace{-10pt}
\label{fig:layout}
\end{figure}

We also implement ViT and the proposed \emam designs using GF 22FDX FDSOI 22~nm technology to obtain detailed area, area, and power comparisons.
Our \emam design employs 8-bit integer precision for both features and weights.
Equivalent quantization was performed for ViT to ensure fair comparisons.
The post-layout results and performance summary of the proposed \emam architecture are shown in Figure \ref{fig:layout}, highlighting several key components.
The core design dimensions are 0.5913 mm $\times$ 0.5917 mm.

The ViT accelerator requires 8,880,532 NAND2-equivalent gates, representing 4.77$\times$ more resources than \emam, which uses only 1,859,065 gates.
As shown in Table~\ref{tab:gate_number_comparison}, \emam consumes 76.10~mW of dynamic power and 95.27~$\mu$W of leakage power at 300~MHz, while ViT consumes 746.95~mW of dynamic power and 3.19~mW of leakage power under the same conditions.
This corresponds to a 9.84$\times$ reduction in total power consumption compared to ViT.
\emam consumes only 1.254~$\mu$J to process the first image, whereas ViT consumes 60.99~$\mu$J.
\emam uses 48.6$\times$ less energy than ViT, mainly due to its 1,643 cycles latency, whereas ViT has 8,130 cycles.
These results highlight the energy efficiency of \emam, making it a strong candidate for edge deployment.

\begin{table}[t]
\centering
\vspace{-4pt}
\caption{Comparison of gate count (NAND2), area, and power between ViT and \emam}
\label{tab:gate_number_comparison}
\begin{tabular}{c|ccc}
\hline \hline
Model & Gate count & Area [mm²] & Power (mW) \\ \hline
ViT   & 8,880,532  & 1.669      & 750.1      \\
\emam & 1,863,608  & 0.350      & 76.20      \\ 
\hline \hline
\end{tabular}
\end{table}

\subsection{Ablation Study}\label{ssec:ablation}

\begin{table}[t]
\caption{Ablation study of accuracy under INT8 on the MARS dataset for Na\"ive Mamba and \emam with approximation techniques adapted from MARCA and I-ViT}
\label{tab:ablation_study}
\centering
\scriptsize
\begin{subtable}[t]{0.49\linewidth}
\centering
\begin{tabular}{cccc|cc}
\hline \hline
\multicolumn{4}{c|}{Ablation}   & \multirow{2}{*}{\begin{tabular}[c]{@{}c@{}}MAE\\ {[}cm{]}\end{tabular}} & \multirow{2}{*}{\begin{tabular}[c]{@{}c@{}}RMSE\\ {[}cm{]}\end{tabular}} \\ \cline{1-4}
\begin{tabular}[c]{@{}c@{}}I-Layer\\ Norm.\end{tabular} & \begin{tabular}[c]{@{}c@{}}MARCA\\ SiLU\end{tabular} & \begin{tabular}[c]{@{}c@{}}Shift\\ Exp.\end{tabular} & \begin{tabular}[c]{@{}c@{}}Piecewise\\ Softplus\end{tabular} &  &   \\ \hline
 - & - & - & - & 8.45 & 10.4 \\
 \checkmark & - & - & - & 26.4 & 29.1 \\
 - & \checkmark & - & - & 8.55 & 10.5 \\
 - & - & \checkmark & - & 13.2 & 19.8 \\
 - & - & - & \checkmark & 8.45 & 10.4 \\ \hline
 \checkmark & \checkmark & \checkmark & \checkmark & 45.5 & 47.9 \\ \hline \hline
\end{tabular}
\caption{Na\"ive Mamba}
\label{tab:ablation_study_naive}
\end{subtable}
\hfill
\begin{subtable}[t]{0.49\linewidth}
\centering
\begin{tabular}{cccc|cc}
\hline \hline
\multicolumn{4}{c|}{Ablation}   & \multirow{2}{*}{\begin{tabular}[c]{@{}c@{}}MAE\\ {[}cm{]}\end{tabular}} & \multirow{2}{*}{\begin{tabular}[c]{@{}c@{}}RMSE\\ {[}cm{]}\end{tabular}} \\ \cline{1-4}
\begin{tabular}[c]{@{}c@{}}Range\\ Norm.\end{tabular} & \begin{tabular}[c]{@{}c@{}}Piecewise\\ SiLU\end{tabular} & \begin{tabular}[c]{@{}c@{}}Piecewise\\ Exp.\end{tabular} & ReLU &  &   \\ \hline
- & - & - & - & 8.45 & 10.4 \\
\checkmark & - & - & - & 6.71 & 8.89 \\
- & \checkmark & - & - & 8.46 & 10.4 \\
- & - & \checkmark & - & 8.46 & 10.4 \\ 
- & - & - & \checkmark & 8.51 & 10.5 \\ \hline
\checkmark & \checkmark & \checkmark & \checkmark & 6.63 & 8.84 \\ \hline \hline
\end{tabular}
\caption{\emam}
\label{tab:ablation_study_emam}
\end{subtable}
\end{table}

Table~\ref{tab:ablation_study} shows ablation study accuracy results under INT8 on the MARS dataset for \textit{na\"ive Mamba} and \emam, focusing on the individual impact of the hardware-oriented optimizations proposed in Section~\ref{sec:design}.
We implement a \textit{na\"ive Mamba} by incorporating the approximation techniques used in I-ViT and MARCA, as exact mathematical implementations are inefficient for hardware.
Since range normalization and ReLU involve structural differences compared to the original \textit{Ref. Mamba}, which does not include these layers, we conducted isolated retraining experiments using only each of these components.
The first row of Table~\ref{tab:ablation_study_naive} and ~\ref{tab:ablation_study_emam} shows the accuracy of \textit{Ref. Mamba} under INT8 precision.
The I-layer normalization employs an improperly estimated scaling factor, resulting in increased errors with an MAE of 26.4 cm and an RMSE of 29.1 cm.
The accuracy degradation is minimal when applying approximations to SiLU and softplus in both na\"ive Mamba and \emam because the piecewise approximation closely resembles the original functions.
In contrast, using ReLU in \emam does not significantly impact accuracy, and range normalization leads to performance improvement, reducing the MAE from 8.45 cm to 6.71 cm and the RMSE from 10.4 to 8.89 cm.
These results confirm that the optimization strategies employed in \emam are more effective and better suited for INT8 quantized environments.

\begin{table}[t]
\centering
\caption{System-level ablation study results analyzing the impact on frame latency and throughput}
\label{tab:ablation_system}
\resizebox{0.9\linewidth}{!}{
\begin{tabular}{cccc|cc}
\hline \hline
 \multicolumn{4}{c|}{Ablation} & \multirow{2}{*}{\shortstack{Frame Latency\\ {[cycles]}}} & \multirow{2}{*}{\shortstack{Throughput\\ {[Mb/s]}}} \\ \cline{1-4}
 Range Norm. & Piecewise SiLU & Piecewise Exp. & ReLU & & \\ \hline
 - & - & - & - & 10,220 & 49.4 \\
 \checkmark & - & - & - & 2,480  & 188  \\
 - & \checkmark & - & - & 10,220 & 49.4 \\
 - & - & \checkmark & - & 10,220 & 49.4 \\
 - & - & - & \checkmark & 10,220 & 49.4 \\ \hline
 \checkmark & \checkmark & \checkmark & \checkmark & 1,643 & 263 \\ \hline \hline
\end{tabular}
}
\end{table}

To systematically evaluate the contribution of each optimization on frame latency and throughput, we present an ablation study in Table~\ref{tab:ablation_system}.
Each approximation is applied individually to the \textit{na\"ive Mamba} baseline to assess its isolated effect on frame latency and throughput on the FPGA implementation.
Among those approximations, only range normalization improves the performance noticeably.
This limited impact of the other techniques arises from the pipelined structure: since normalization is the first operation in the Mamba block and the I-layer normalization from I-ViT, as used in \textit{na\"ive Mamba}, has high latency, subsequent layers cannot begin computation until normalization is complete.
The piecewise approximation of SiLU and replacing ReLU do not affect the performance since they have the same latency as the SiLU in MARCA and the piecewise softplus.
When all optimizations are applied together, each optimization reduces latency, resulting in significant system-level performance gains.
In particular, the execution time of range normalization is substantially reduced compared to the I-layer normalization used in \textit{na\"ive Mamba}, alleviating the pipeline bottleneck.
As a result, the latency improvements introduced by subsequent layers, such as the exponential functions, can take effect earlier and propagate through the pipeline, allowing the benefits of all optimizations to manifest at the system level.
\section{Conclusions}~\label{sec:conclusion}
Mamba, a recently proposed SSM-based DL architecture, has garnered significant attention for achieving competitive accuracy compared to state-of-the-art models while offering higher processing efficiency due to its linear complexity. 
Despite its potential, no hardware accelerator design framework has targeted Mamba for resource-constrained edge AI applications.
This paper presented \emam, a comprehensive end-to-end hardware acceleration framework designed explicitly for deploying Mamba in edge environments.
\emam introduces hardware and application-aware approximations for computationally expensive non-linear operations such as SiLU activation and exponentiation.
Furthermore, it incorporates an efficient accelerator architecture tailored to SSM with quantization and employs a NAS to optimize hyperparameters for resource-constrained environments.
Evaluations with multiple edge applications, including Fashion-MNIST, CIFAR-10, and MARS datasets, show that \emam achieves accuracy comparable to state-of-the-art techniques \textit{using 1.63–19.9$\times$ fewer parameters}.
In addition, \emam generalizes well to large-scale natural language tasks, demonstrating stable perplexity across varying sequence lengths on the WikiText2 dataset.
We also implement \emam on AMD ZCU102 FPGA and ASIC using GF 22~nm technology and compare it against CNN and ViT baselines.
The accuracy and performance impact of each component within the NAS process is analyzed through an ablation study.
\emam achieves 5.62$\times$ lower latency and 9.95× higher throughput using 63$\times$ fewer parameters compared to the CNN baseline. 
When compared to the ViT baseline, it achieves 4.95$\times$ lower latency and 2.22$\times$ higher throughput using 1.63$\times$ fewer parameters and 4.77$\times$ smaller area, along with 9.84$\times$ and 48.6$\times$ lower power and energy consumption, respectively. These results validate \emam’s effectiveness and scalability for real-world edge AI deployments.

\section{Acknowledgments}~\label{sec:acknowledgments}
This work was supported by the MSIT, Korea, under the Global Research Support Program in the Digital Field program(No. RS-2024-00431397) supervised by the IITP and by the National Research Foundation of Korea grant funded by the Korea government(MSIT) (No. RS-2023-00208046).

\bibliographystyle{ACM-Reference-Format}
\bibliography{References/accelerator,References/mri,References/ssms,References/model}

\end{document}